\documentclass[journal]{IEEEtran}
\usepackage{graphicx}

\input{tcilatex}

\begin{document}

\title{Multi-task UNet architecture for end-to-end autonomous driving}

\author{Der-Hau~Lee and~Jinn-Liang~Liu 
\thanks{%
This work was supported by the Ministry of Science and Technology, Taiwan,
under Grant MOST 109-2115-M-007-011-MY2.} 
\thanks{%
D. Lee was with the Department of Electrophysics, National Chiao Tung
University, Hsinchu 300, Taiwan.} 
\thanks{%
J. Liu is with the Institute of Computational and Modeling Science, National
Tsing Hua University, Hsinchu 300, Taiwan (e-mail:
jinnliu@mail.nd.nthu.edu.tw, website: http://www.nhcue.edu.tw/~jinnliu).} }

\maketitle

\begin{abstract}
We propose an end-to-end driving model that integrates a multi-task UNet (MTUNet) architecture and control algorithms in a pipeline of data flow from a front camera through this model to driving decisions. It provides quantitative measures to evaluate the holistic, dynamic, and real-time performance of end-to-end driving systems and thus the safety and interpretability of MTUNet. The architecture consists of one segmentation, one regression, and two classification tasks for lane segmentation, path prediction, and vehicle controls. We present three variants of the architecture having different complexities, compare them on different tasks in four static measures for both single and multiple tasks, and then identify the best one by two additional dynamic measures in real-time simulation. Our results show that the performance of the proposed supervised learning model is comparable to that of a reinforcement learning model on curvy roads for the same task, which is not end-to-end but multi-module. 
\end{abstract}

\section{Introduction}

End-to-end driving system with a single deep neural network (DNN) is an emerging technology in autonomous vehicles \cite{Gri20,Tam20,Yur20,Bad21}. The system is a pipeline consisting of perception sensors, DNN, and control actuators \cite{Gri20,Tam20,Yur20} with a data flow from sensors to DNN, path planning, controllers, and then to actuators for making driving decisions of steering, acceleration, or braking in an end-to-end, autonomous, and real-time manner \cite{Gri20,Tam20,Yur20,Lee21}.

Since Pomerleau's pioneering work in the 1980s \cite{Pom89}, a variety of end-to-end DNNs have been proposed for various tasks in autonomous driving \cite{Lee21,Lec06,Huv15,Che15,Xu17,Sau18,Tei18,Cud20,Qia20,Lee21a}.
Most of these DNNs belong to single-task learning models having single (regression or probabilistic) loss function for training the model to infer single driving task (steering angle, lead car's distance, or turning etc.). 

Autonomous vehicles are equipped with various sensors (cameras, LiDARs, Radars, GPS etc.) to tackle complex driving problems (localization, object detection, scene semantics, path planning, maneuvers etc.) \cite{Fen20,Van21}. Multi-task (MT) deep learning can usually achieve better performance than its single-task (ST) counterparts due to more data from different tasks \cite{Zha21}. However, it remains a challenge to design MTDNNs that use fused multi-modal data to achieve stringent requirements of accuracy, robustness, and real-time performance in autonomous driving \cite{Fen20,Van21}.

There is another issue in end-to-end driving pipeline, namely, the integral system of DNN, path planning \cite{Bad21,Cui19}, and control \cite{Lee21,Cud20,Qiu15,Li19,Dea20,Thr06,Sam21} algorithms that meets safety and comfort requirements. These algorithms are generally proposed and verified separately since automotive control systems are very complex varying with vehicle types and levels of automation \cite{Gri20,Vah03,Bag16,Hus19}. The literature is very scarce on the overall and dynamic evaluation of these algorithms in an integrated pipeline for multiple tasks using multi-modal data \cite{Lee21,Sau18,Tei18}.

We propose here a multi-task deep learning model based on the well-known UNet architecture of image semantic segmentation \cite{Ron15,Lon15,Meh18,Zha18}. Our MTUNet can infer one segmentation, one regression, and two classification tasks for a vehicle to perform lane segmentation, estimate its heading angle, and classify its road path ahead and its distance to lead cars, respectively.  We reduce 5 indicators in our previous work \cite{Lee21} to 2, change one overtaking classification in \cite{Lee21} to two pose classifications, and add a lane semantic segmentation \cite{Lee21a} in the present work. The driving tasks are different between these two works. The coupled approach of lateral and longitudinal controllers in \cite{Lee21} aims at arbitrary lane changing and overtaking and thus requires to incorporate multiple sensors to avoid collisions in highway traffic \cite{Lee21}. We use here a decoupled approach with a modified lateral and a   longitudinal controller, which is mainly for lane keeping, a primary function  of  active driving assistance systems currently being developed and deployed to commercial vehicles for Level 2 autonomous driving \cite{Con20}.

\section{Learning and control models}

\subsection{Multi-task UNets}
    
Figure 1 illustrates the architecture of MTUNets with MTResUNet \cite{Lee21a}, a residual UNet \cite{Zha18} for example.

\textbf{Backbone, segmentation, and pose subnets.} UNets have backbone (blue encoder in Fig. 1) and segmentation (black decoder) subnets for lane segmentation task. We refer to \cite{Lee21a} for more details about the description and comparison of various UNets including MTUNet (extended directly from the original UNet \cite{Ron15}), MTResUNet, and MTDSUnet used in the present work, where DS is short for depthwise separable. The pose (red) subnet connects to the last layer of the backbone and is designed to perform a regression task (the top branch of the subnet in Fig. 1) and two classification tasks (the other two branches). A total of four tasks considered here instead of a single segmentation task in \cite{Lee21a}.

MTUNet and MTDSUNet have similar structures as shown in Fig. 1 except that MTUNet does not have skip connections, and MTDSUNet replaces the standard convolutions of MTResUNet by depthwise separable convolutions to improve computational complexity \cite{Lee21a,Cho17}. Table 1 shows the complexity of these three models in the total number of parameters (Params), multiply-and-accumulates (MACs), and frames per second (FPS) in inference speed. MTDSUNet is 2.26$\times$ lighter and 1.78$\times$ faster than MTResUNet in Params and FPS, respectively. 

\begin{figure}[t]
\centerline{\includegraphics[scale=0.36]{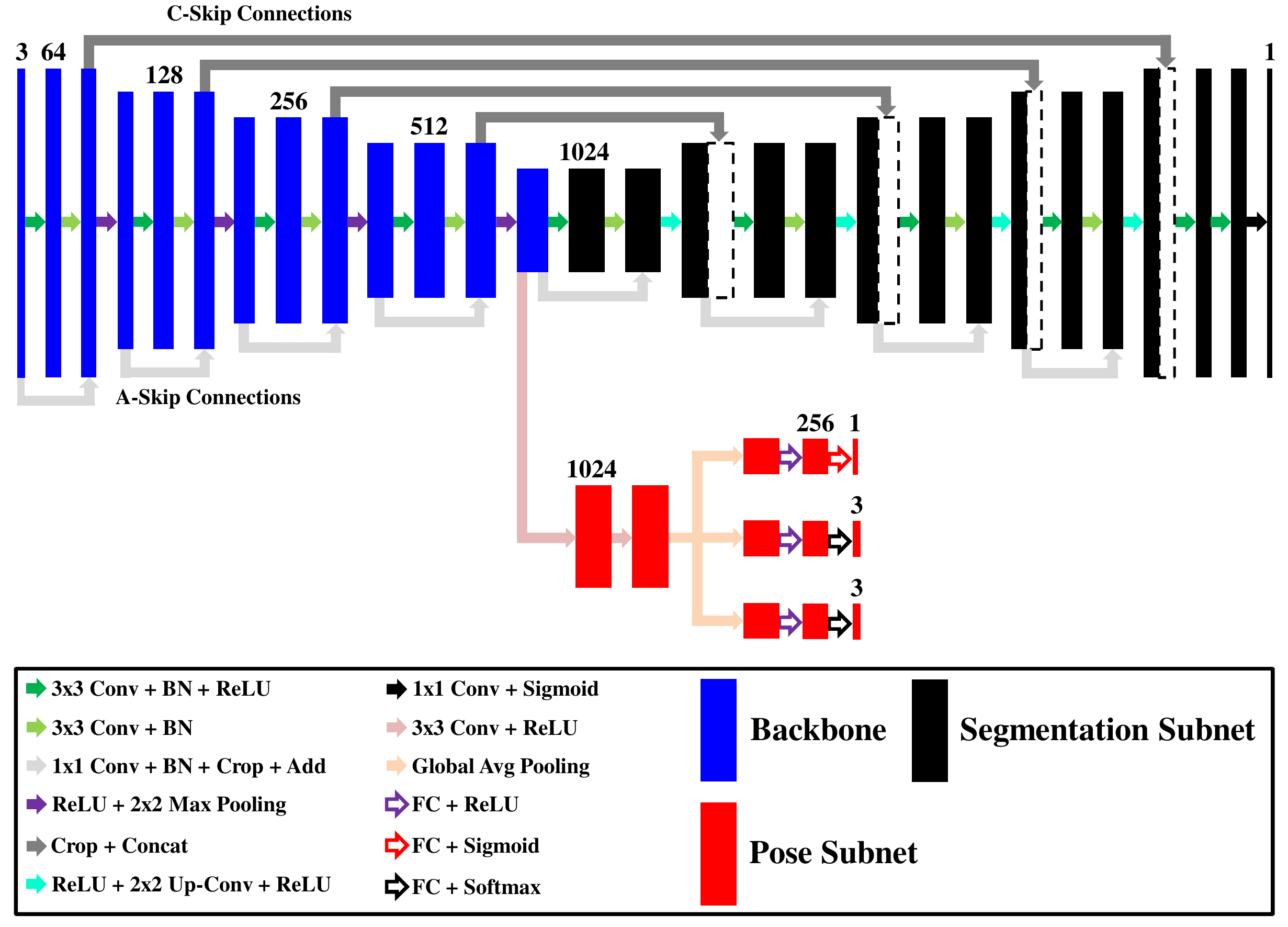}}
\caption{Multi-task UNet architecture with ResUNet for example. }
\end{figure}

\begin{table}
\caption{Complexity of MTUNets in parameters (Params), multiply-and-accumulates (MACs), and frames per second (FPS)}
\begin{center}
\begin{tabular}{lccc}
\hline
  & Params & MACs & FPS \\ \hline
MTResUNet   &  47.38 M & 70.37 B &  23  \\ 
MTUNet   & 45.99 M & 67.46 B & 26  \\
MTDSUNet & {\bf 20.96} M  & {\bf 14.50} B & {\bf 41}  \\ \hline
\end{tabular}
\end{center}
\end{table}

The regression task estimates Ego's heading angle while the classification tasks infer the road type (left turn, straight, or right turn) ahead Ego and LCar's distance (far, nearby, or close) from Ego, if any. These tasks are defined more precisely as follows.

The output of the segmentation subnet has the same resolution of the input image and is transformed to a single channel of a gray (probability) image of lane lines by a pixel-wise sigmoid operation \cite{Lee21a}. The normalized cross entropy loss \cite{Lee21a}
\begin{eqnarray}    
L_S  =  -\frac{N}{P + N} \sum\limits_{i = 1, {\tilde x}_i  = 1}^P \log \left( \sigma(x_i) \right)
\notag \\
- \frac{P}{P + N} \sum\limits_{i = 1, {\tilde x}_i  = 0}^N \log \left( 1 - \sigma(x_i) \right) 
\end{eqnarray}
is used to train MTUNets with the other three tasks, where $x_i \in (0,1)$ is a predicted score, ${\tilde x}_i = 0$ or 1 is the corresponding ground-truth value, $P$ and $N$ are the total numbers of positive (lane line) and negative (background) pixels in a batch of ground-truth images, respectively, and $\sigma $ is the sigmoid function. 

The segmentation subnet can thus generate a sequence of gray images of predicted lane lines with estimated lane width and lane centerline \cite{Lee21a}, which will be used in path prediction.

We have used two CNNs in \cite{Lee21} to estimate LCar's distance by regression approach which yielded errors larger than 3 meters, far greater than that by radars (less than 10 cm \cite{Goh11}). However, radars cannot perceive some obstacle categories that CNNs can with camera's images. Therefore, we adopt classification approach here to perceive and categorize LCar into three classes in distance so that the classification output can be integrated with radar's distance output into control algorithms \cite{Lee21} for autonomous driving.

The pose subnet consists of two convolutional layers shared by three parallel global average pooling (GAP) layers that extract features for the corresponding three tasks at low spatial resolution. This branched NN structure is easier to interpret and less prone to overfitting \cite{Lin13}. Each branch has two fully connected (FC) layers followed by an activation (ReLU, sigmoid, or softmax) layer. Dropout applies to GAP and first FC layers to further prevent overfitting. The loss function of the regression task is a Euclidean L2 norm defined as

\begin{equation}
L_R = \frac{1}{2M}\sum\limits_{i = 1}^M \left\vert \tilde{\theta}_i  - \theta_i \right\vert^2,
\end{equation}
where $\tilde{\theta}$ and $\theta$ are normalized true and estimated values of Ego's heading angle, respectively, and $M$ is the batch size of input images. The cross-entropy

\begin{equation}
L_{C1,C2} = \frac{-1}{M}\sum\limits_{i = 1}^M \sum\limits_{j = 1}^3 \tilde p_{ij} \log(p_{ij})
\end{equation}
is for two classification tasks (C1 and C2), where $\tilde p$ and $p$ are true and sigmoid (C1) (or softmax (C2)) values of three classes, respectively. The three classes of C1 are denoted by C1L, C1S, and C1R for left turn, straight, and right turn, respectively, and C2F, C2N, and C2C of C2 for LCar's far, nearby, and close distance.

The pose subnet does not change the topology of U-shaped network and can output multiple road predictions and pose estimations in a single forward pass, which are crucial in self-driving cars to make driving decisions \cite{Meh18}. Moreover, the combination of regression and classification losses yields a multi-modal loss that can avoid mode collapse problems when each individual task is trained alone \cite{Xu17,Fen20,Cui19}.

In \cite{Lee21a}, we proposed a CNN-based path prediction (PP) algorithm that uses advanced methods to deal with complexity, scalability, and homography issues of CNN's output images. Ego’s heading angle $\theta$ and lateral offset $\Delta$ to lane center are outputs of MTUNets and PP, respectively \cite{Lee21a}. Since MTUNets learn from road images to predict $\theta$, it is important to feed annotated images to MTUNets with different classes like C1L, C1S, and C1R that teach MTUNets how to perceive curvy road with meaningful $\theta$, i.e., without mode collapsing. For longitudinal motion, we design another three classes C2F,  C2N, and  C2C for MTUNets to learn how to roughly discern LCar's distance in case that radar fails to detect LCar.

\subsection{Control}

We define the steering command
\begin{equation}
{\rm SteerCmd}(C, \theta, \Delta, v) = c_1 S_t 
\end{equation} 
in discrete time $t$ as a function of Ego's driving desire $C$ (= 0 for staying in current, -1 changing to left, or 1 to right lane), $\theta$ from MTUNets, $\Delta$ from PP, and Ego's speed $v$ by using the Stanley controller 
\begin{eqnarray}
S_t  &=& \bar{S}_t  - c_2 ( \bar{S}_t  - S_{t-1}) \\
\bar{S}_t  &=& \theta_t  + \arctan \left( {c_3 \frac{{\Delta_t  + C W}}{{v_t }}} \right), 
\end{eqnarray}
which is a geometric trajectory tracking control model \cite{Sau18,Thr06,Sam21}.
Here, $c_1$ is a constant normalizing the steering value to the range [-1,1], $c_2 = 0.5$ a damping parameter, $c_3 = 2.5$ a gain parameter, and $W=4$ m a lane width (it can be a variable from CNN's lane detection). The Stanley controller iteratively adjusts the steering angle to bring Ego to lane center, and is shown to improve the robustness of a learning based planner recently \cite{Fan22}. The present work focuses on lane-keeping control.

The discrete time PI controller \cite{Sam21}
\begin{equation}
{\rm PI}(v) = {u} = k_p e_t  + k_I \sum\limits_{i = 1}^n e_i \Delta \tau
\end{equation}
is used to control Ego in longitudinal motion, where $u$ is control action (acceleration or deceleration or none) at current time $t$, $e_t=v_t-v_r$ is the error between Ego's current ($v_t$) and reference ($v_r$) speed, $n$ is the number of sampling instances, $\Delta \tau$ is the time between instances, and $k_p = 2$ and $k_I = 0.5$ are proportional and integral gains. We maintain  Ego's speed to  $v_r$ during entire lane-keeping maneuver, which is more  challenging  than traditional urban driving, i.e., Ego decelerates before entering a corner. In addition, Ego with this driving behavior is possible to reduce lap time for high-speed autonomous racing. The action value is then normalized to
\begin{equation}
{\rm AccelCmd} = \tanh \left( u \right)
\end{equation}
by the tanh function so is SteerCmd. 

These control models are thus combined with MTUNets to form an MTUC algorithm, which is then implemented in the real-time simulator developed in \cite{Lee21a} for evaluation.

\section{Experiment setup}
To assess the MTUC algorithm in static and dynamic conditions \cite{Lee21}, we prepare and annotate input data for training and testing as follows. 

\begin{figure}[!t]
\centerline{\includegraphics[scale=0.33]{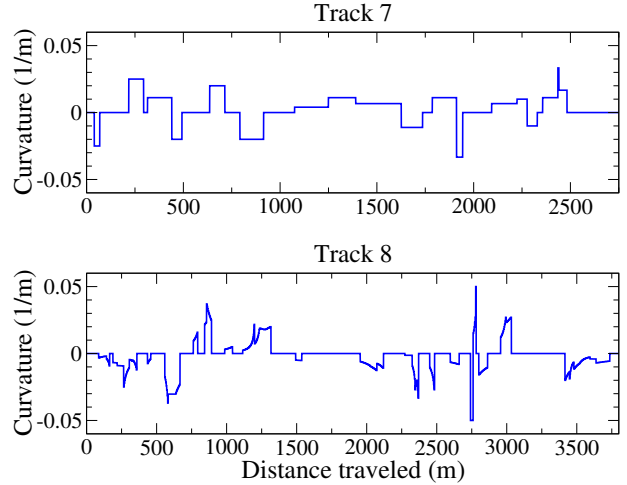}}
\caption{Curvature profiles of Tracks 7 and 8 \cite{Che15}, where the lane width is 4 m, the total lengths are 2843 and 3919 m, respectively. Track 8 is curvier than Track 7, for example, the shape curves near 2800 m represent sudden changes in the road direction.}
\end{figure}

\textbf{Image data.}
We use the method in \cite{Lee21} to collect additional image data for this work with a modification. Ego drives normally with other agents on six different tracks \cite{Che15} (not shown) to collect 39,185 and 3,562 images (42,747 in total) for training and testing, respectively. It is normal instead of zigzag \cite{Lee21} driving because the number of affordance indicators reduces from 5 in \cite{Lee21} to 2 here since MTUC is mainly for studying MTUNets and controllers in lane keeping maneuver instead of autonomous lane changing and overtaking \cite{Lee21}. Another two curvy tracks, namely, Tracks 7 and 8 in Fig. 2, are used to evaluate MTUC’s dynamic performance in unseen environments.

\textbf{Labeling multi-modal data.}
For road-type classification task, each image is labeled to one of three classes C1L, C1S, and C1R  according to its corresponding value of Ego's angle, where C1L, C1S, and C1R represent $\tilde{\theta} < -0.006$ (left turn), $-0.006 \le \tilde{\theta} \le 0.006$ (straight), and $0.006 < \tilde{\theta}$ (right turn) in radian, respectively.

We annotate each image (at $640 \times 480$ resolution) with bounding boxes to other cars and label the closest box as C2F, C2N, or C2C for boxA = 0 (far), 0 $<$ boxA $<$ 3200 (nearby), or 3200 $\le$ boxA (close), where boxA denotes the area of the box in pixel$^2$. The image is also processed to obtain a binary (black and white) image of lane-line masks as the ground-truth of lane segmentation for MTUNets. All binary images are similar to those annotated from real images \cite{Ko21}. Numbers of labeled multi-modal images  of (C1L, C1S, C1R) and (C2F, C2N, C2C) classes are (8461, 26179, 8107) and (16405, 11311, 15031), respectively.

\textbf{Training strategy.}
We first train the pose subnet by stochastic gradient descent with the batch size bs = 20, momentum m = 0.9, and learning rate starting from lr = 0.01 and decreasing by a factor of 0.9 every 5 epochs for a total of 100 epochs. 

Segmentation and pose subnets are then trained jointly using Adam optimizer with bs = 1, m = 0.9, and lr = $10^{ - 4}$ and $10^{ - 5}$ for first 75 and last 25 epochs, respectively. Segmentation subnet is also trained independently for a comparison between single and multi task cases. The total loss in each stage is a weighted sum of the corresponding losses in Eqs. (1) - (3).

\textbf{Performance measures.} Lane segmentation results from the segmentation subnet of MTUNets in Fig. 1 are expressed in pixel-wise true
positive, true negative, false positive, and false negative
values which are paired with ground truth values of lane lines in images.  MTUNets'
performance is measured in accuracy, precision, recall, and F1 score \cite{Huv15}.
For the pose subnet, mean absolute error (MAE) and Accuracy are used to evaluate regression (Heading for $\theta$) and classification (Road Type for C1 and LCar Dist. for C2) tasks, respectively.

\section{Results and discussions}

Table 2 shows that the performance of UNets is in the reverse order of their complexity in Table 1 as expected, i.e., the higher complexity of the model yields better measures but worse efficiency (FPS). Table 2 also shows that single-task (without MT) UNets outperform their multi-task (with MT) counterparts since the correlated tasks in the MT case obviously incur more errors than individual tasks. Nevertheless, differences between these two cases are very small in all measures. We exclude the worst DSUNet in Table 2 from what follows for further investigation on the dynamic performance of MTUC when Ego drives along with other cars on Track 7/8 in Fig. 2. 

\begin{table}
\caption{Performance of trained UNets for single (without MT) and multiple (MT) tasks on test data for (a) segmentation and (b) pose tasks}
\begin{center}
\begin{tabular}{ccccc}
\hline
(a) & Accuracy & Precision & Recall & F1 Score \\ \hline
DSUNet & {\bf 0.990} & {\bf 0.902} & {\bf 0.833} & {\bf 0.858} \\
MT & 0.988 & 0.837 & 0.831 & 0.828 \\ \hline
UNet & {\bf 0.995} & {\bf 0.933} & {\bf 0.911} & {\bf 0.921} \\
MT & 0.992 & 0.883 & 0.899 & 0.889 \\ \hline
ResUNet & {\bf 0.995} & {\bf 0.937} & {\bf 0.912} & {\bf 0.923} \\
MT & 0.993 & 0.898 & 0.893 & 0.893 \\
\hline
 &  &  &  &  \\
\end{tabular}
\begin{tabular}{cccc}
\hline
(b) & Heading & Road Type & LCar Dist. \\ 
& MAE & Accuracy & Accuracy \\ \hline
DSUNet & 0.009 & {\bf 0.723} &  0.941 \\
MT & {\bf 0.007} &  0.681 & {\bf 0.942} \\ \hline
UNet & 0.004 & 0.970 & {\bf 0.952} \\
MT & 0.004 & 0.970& 0.949 \\ \hline
ResUNet & 0.004 & {\bf 0.969} & {\bf 0.956} \\
MT & 0.004 & 0.968 & 0.954 \\
\hline
\end{tabular}
\end{center}
\end{table}

MTUC accounts for Ego's variable speed (up to 76 km/h) and physical properties (weight 1150 kg, length 4.52 m, and width 1.94 m \cite{Wym00,Mun10}) to prevent sliding, slipping, and rollover in curves \cite{Par15}. Table 3 and Fig. 3/4 show MTUC's dynamic performance using MTUNet and MTResUNet in real-time simulation at maximum speed 76/50 km/h on Track 7/8, where the dynamic mean absolute error (dMAE) \cite{Lee21} of predicted angle $\theta$ (blue curves in radian) and dMA lateral offset $\Delta$ (blue curves in m) are calculated while Ego is in motion. The ground-truth values (red curves) of $\theta$ and $\Delta$ are given by the open
racing car simulator (TORCS) \cite{Wym00,Mun10} and set to zero, respectively.

\begin{table}
\caption{Performance of MTUC algorithm with MTUNets in the dynamic mean absolute error (dMAE) of Ego's heading angle $\theta$ and dMA lateral offset $\Delta$ on Track 7/8 in Fig. 3/4}
\begin{center}
\begin{tabular}{lcccc}
\hline
  & \multicolumn{2}{c}{Track 7} & \multicolumn{2}{c}{Track 8} \\ 
  & $\theta$ (rad.) & $\Delta $ (cm) & $\theta$ (rad.) & $\Delta $ (cm) \\ \hline
MTUNet    & {\bf 0.0109} & 14.18  & {\bf 0.0058} & {\bf 6.33} \\ 
MTResUNet & 0.0175 & {\bf 12.80} & 0.0059 & 6.58 \\ 
\hline
\end{tabular}
\end{center}
\end{table}

The lateral indicators $\theta$ and $\Delta$ are crucial measures to quantify and interpret dynamic effects of CNN's perception on path planning in autonomous driving. MTUC with MTResUNet performs better than that of \cite{Cud20} (cf. Table 3), i.e., 0.0175/0.0059 vs. 0.029 and 12.8/6.5 vs. 45.3 cm in $\theta$-dMAE and dMA $\Delta$, respectively, in centered driving style on different (artificial vs. real-world) roads by different (artificial vs. real) cars at different maximum speeds (76/50 vs. 54 km/h). The end-to-end driving algorithm in \cite{Cud20} includes $\Delta$ calculations but without using learning methods for lateral corrections like our lateral offset estimation. The experimental roads in \cite{Cud20} are less curvy than Track 7/8. The minimum radius of general roads is about 130-160 m \cite{Fit94} corresponding to a maximum curvature of 0.00625-0.0076 whereas, for example, that of Track 7 is approximately 0.03 ${\rm m}^{-1}$. Table 3 and Fig. 3/4 thus postulate that MTUC may reduce dMA $\Delta$ to within 10 cm on real roads in future studies. 

\begin{figure}[!t]
\centerline{\includegraphics[scale=0.33]{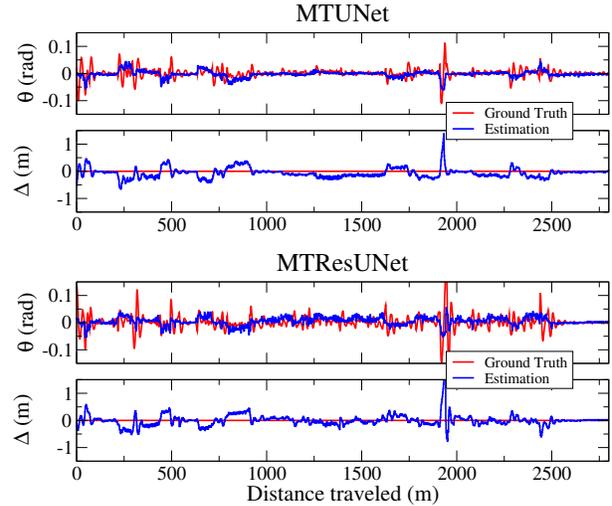}}
\caption{Dynamic performance of MTUC algorithm with MTUNets in $\theta$-dMAE and dMA $\Delta$ for Ego's lane keeping maneuver along Track 7}
\end{figure}

\begin{figure}[!t]
\centerline{\includegraphics[scale=0.33]{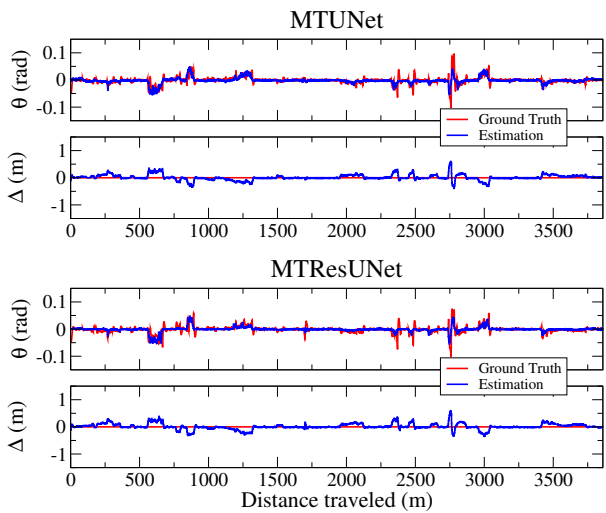}}
\caption{Dynamic performance of MTUC algorithm with MTUNets in $\theta$-dMAE and dMA $\Delta$ for Ego's lane keeping maneuver along Track 8}
\end{figure}

Li et al. investigated the lateral control of an MTCNN with 5 Conv layers and an MT reinforcement learning (MTRL) model using TORCS and a multi-module method (perception and control modules) instead of end-to-end learning method \cite{Li19}. They showed that MTRL outperforms MTCNN and yields 0.01 and 14.8 cm in $\theta$-dMAE and dMA $\Delta$, respectively, on a similar track. Our results are comparable to theirs as shown in Table 3. Figs. 3 and 4 thus show that the end-to-end MTUC algorithm effectively performs lateral control maneuvers in terms of quantitative measures under dynamic, real-time, and variable driving conditions. The qualitative performance of MTUC with ResUNet for inferring segmentation, road types, LCar's distance, and Ego's heading is illustrated Fig. 5. 

\begin{figure}
\centerline{\includegraphics[scale=0.13]{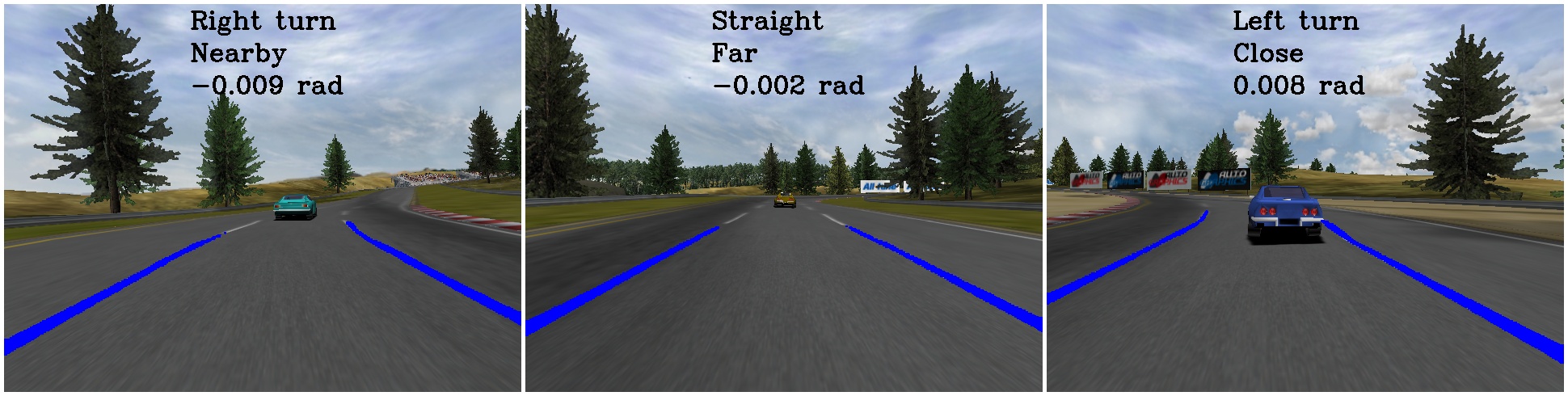}}
\caption{Qualitative performance of MTUC with ResUNet for inferring segmentation (blue lines), road types (Right turn, Straight, and Left turn), LCar's distance (Nearby, Far, Close), and Ego's heading (-0.009, -0.002, and 0.008 rad.) }
\end{figure}

\section{Conclusion}
We proposed a multi-task Unet based control (MTUC) algorithm that combines multi-task UNet, path prediction, and control models. MTUC uses camera images and ego car's pose information as input data to perform end-to-end autonomous driving. Three UNet variants are firstly compared in complexity and four performance measures, from which two are then compared in real-time simulation using dynamic measures of heading angle and lateral offset. Finally, the best one is chosen for further testing on lane keeping maneuver and car following. These measures are important to evaluate the safety and interpretability of end-to-end driving systems with single deep learning neural network (DNN). It is shown that the proposed self-driving algorithm improves our  previous approach \cite{Lee21}, and is comparable to a reinforcement learning model, which is not end-to-end but multi-module \cite{Li19}. 


\end{document}